\documentclass{vgtc}                          

\graphicspath{{figures/}{pictures/}{images/}{./}} 

\usepackage{times}                     

\usepackage{tabu}                      
\usepackage{booktabs}                  
\usepackage{lipsum}                    
\usepackage{mwe}                       

\usepackage{mathptmx}                  

\usepackage{times}
\usepackage{xspace}
\usepackage{multirow}

\usepackage{subcaption}
\usepackage{url} 

\usepackage{colortbl}
\definecolor{green}{rgb}{0.1,0.1,0.1}

\def\approach{SCGS\xspace}
\usepackage[nolist]{acronym}

\vgtcinsertpkg

\title{Semantics-Controlled Gaussian Splatting for Outdoor Scene \\Reconstruction and Rendering in Virtual Reality}

\author{Hannah Schieber\thanks{e-mail: hannah.schieber@tum.de}
\and Jacob Young\thanks{e-mail: jacob.young@otago.ac.nz}
\and Tobias Langlotz\thanks{e-mail: tobias.langlotz@otago.ac.nz}
\and Stefanie Zollmann\thanks{e-mail: stefanie.zollmann@otago.ac.nz}
\and Daniel Roth\thanks{e-mail: daniel.roth@tum.de} \\ 
\and \parbox{2.0in}{\scriptsize \centering Department of Artificial Intelligence in Biomedical Engineering\\
Friedrich-Alexander-Universität Erlangen-Nürnberg (FAU) \\ Erlangen, Germany$^{\textsc{*}}$} 
\and \parbox{2.2in}{\scriptsize \centering Department of Computer Science, \\ University of Otago, \\ Dunedin, New Zealand$^{\textsc{†, ‡,§}}$}
\parbox{2.0in}{\scriptsize \centering Technical University of Munich  \\
Human-Centered Computing and Extended Reality Lab  \\
TUM University Hospital \\
Orthopedics and Sports Orthopedics$^{\textsc{*,¶}}$}
}

\teaser{
 \includegraphics[width=\textwidth]{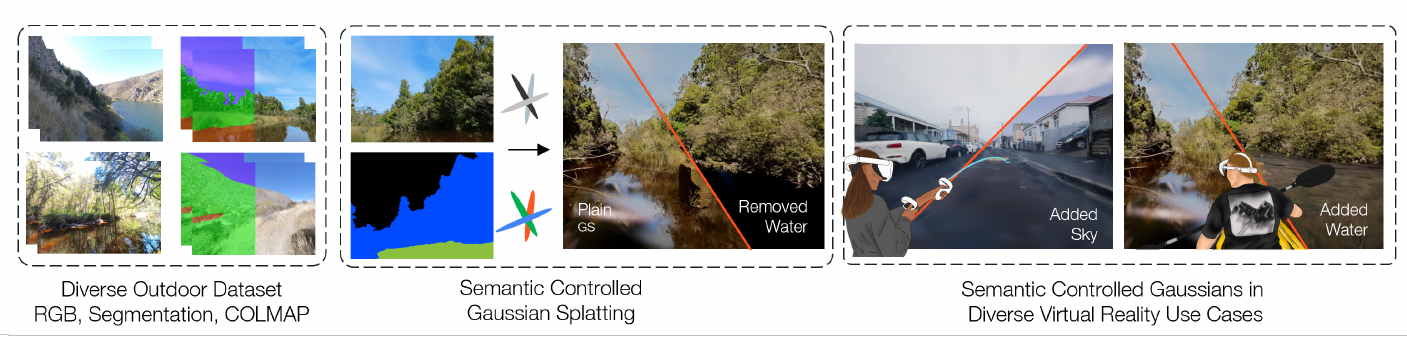}
 \caption{\textbf{\approach generates large-scale 3D assets for a variety of virtual reality applications.} Our approach enables experiencing a virtual environment captured by a continuous camera stream using a \ac{gs} representation. Using semantic segmentation we can replace Gaussian Splats of different classes, e.g., ``water'' for more interactive experiences.}
  \label{fig:teaser}
}

\abstract{

Advancements in 3D rendering like Gaussian Splatting (GS) allow novel view synthesis and real-time rendering in virtual reality (VR). However, GS-created 3D environments are often difficult to edit. For scene enhancement or to incorporate 3D assets, segmenting Gaussians by class is essential.
Existing segmentation approaches are typically limited to certain types of scenes, e.g., ``circular'' scenes, to determine clear object boundaries. 
However, this method is ineffective when removing large objects in non-``circling'' scenes such as large outdoor scenes.

We propose Semantics-Controlled GS (SCGS), a segmentation-driven GS approach, enabling the separation of large scene parts in uncontrolled, natural environments. SCGS allows scene editing and the extraction of scene parts for VR. Additionally, we introduce a challenging outdoor dataset, overcoming the ``circling" setup. We outperform the state-of-the-art in visual quality on our dataset and in segmentation quality on the 3D-OVS dataset. We conducted an exploratory user study, comparing a 360-video, plain GS, and SCGS in VR with a fixed viewpoint. In our subsequent main study, users were allowed to move freely, evaluating plain GS and SCGS. Our main study results show that participants clearly prefer SCGS over plain GS. We overall present an innovative approach that surpasses the state-of-the-art both technically and in user experience.

}

\begin{document}
\begin{acronym}[Bspwwww.]  

\acro{ar}[AR]{augmented reality}
\acro{ap}[AP]{average precision}
\acro{api}[API]{application programming interface}
\acroplural{ann}[ANN]{artifical neural networks}
\acro{bev}[BEV]{bird eye view}
\acro{rbob}[BRB]{Bottleneck residual block}
\acroplural{rbob}[BRBs]{Bottleneck residual blocks}
\acro{mbiou}[mBIoU]{mean Boundary Intersection over Union}
\acro{cai}[CAI]{computer-assisted intervention}
\acro{ce}[CE]{cross entropy}
\acro{cad}[CAD]{computer-aided design}
\acro{cnn}[CNN]{convolutional neural network}

\acro{crf}[CRF]{conditional random fields}
\acro{dpc}[DPC]{dense prediction cells}
\acro{dla}[DLA]{deep layer aggregation}
\acro{dnn}[DNN]{deep neural network}
\acroplural{dnn}[DNNs]{deep neural networks}

\acro{da}[DA]{domain adaption}
\acro{dr}[DR]{domain randomization}
\acro{fat}[FAT]{falling things}
\acro{fcn}[FCN]{fully convolutional network}
\acroplural{fcn}[FCNs]{fully convolutional networks}
\acro{fov}[FoV]{field of view}
\acro{fv}[FV]{front view}
\acro{fp}[FP]{False Positive}
\acro{fpn}[FPN]{feature Pyramid network}
\acro{fn}[FN]{False Negative}
\acro{fmss}[FMSS]{fast motion sickness scale}
\acro{gan}[GAN]{generative adversarial network}
\acroplural{gan}[GANs]{generative adversarial networks}
\acro{gcn}[GCN]{graph convolutional network}
\acroplural{gcn}[GCNs]{graph convolutional networks}
\acro{gs}[GS]{Gaussian Splatting}
\acro{hmi}[HMI]{Human-Machine-Interaction}
\acro{hmd}[HMD]{Head Mounted Display}
\acroplural{hmd}[HMDs]{head mounted displays}
\acro{iou}[IoU]{intersection over union}
\acro{irb}[IRB]{inverted residual bock}
\acroplural{irb}[IRBs]{inverted residual blocks}
\acro{ipq}[IPQ]{igroup presence questionnaire}
\acro{knn}[KNN]{k-nearest-neighbor}
\acro{lidar}[LiDAR]{light detection and ranging}
\acro{lsfe}[LSFE]{large scale feature extractor}
\acro{llm}[LLM]{large language model}
\acro{map}[mAP]{mean average precision}
\acro{mc}[MC]{mismatch correction module}
\acro{miou}[mIoU]{mean intersection over union}
\acro{mis}[MIS]{Minimally Invasive Surgery}
\acro{msdl}[MSDL]{Multi-Scale Dice Loss}
\acro{ml}[ML]{Machine Learning}
\acro{mlp}[MLP]{multilayer perception}
\acro{miou}[mIoU]{mean Intersection over Union}
\acro{nn}[NN]{neural network}
\acroplural{nn}[NNs]{neural networks}
\acro{ndd}[NDDS]{NVIDIA Deep Learning Data Synthesizer}
\acro{nocs}[NOCS]{Normalized Object Coordiante Space}
\acro{nerf}[NeRF]{Neural Radiance Fields}
\acro{NVISII}[NVISII]{NVIDIA Scene Imaging Interface}
\acro{ngp}[NGP]{Neural Graphics Primitives}
\acro{or}[OR]{Operating Room}
\acro{pbr}[PBR]{physically based rendering}
\acro{psnr}[PSNR]{peak signal-to-noise ratio}
\acro{pnp}[PnP]{Perspective-n-Point}
\acro{rv}[RV]{range view}
\acro{roi}[ROI]{region of interest}
\acroplural{roi}[ROIs]{region of interests}
\acro{rbab}[BB]{residual basic block}
\acro{ras}[RAS]{robot-assisted surgery}
\acroplural{rbab}[BBs]{residual basic blocks}
\acro{spp}[SPP]{spatial pyramid pooling}
\acro{sh}[SH]{spherical harmonics}
\acro{sgd}[SGD]{stochastic gradient descent}
\acro{sdf}[SDF]{signed distance field}
\acro{sfm}[SfM]{structure-from-motion}
\acro{sam}[SAM]{Segment-Anything}
\acro{sus}[SUS]{system usability scale}
\acro{ssim}[SSIM]{structural similarity index measure}
\acro{sfm}[SfM]{structure from motion}
\acro{slam}[SLAM]{simultaneous localization and mapping}
\acro{tp}[TP]{True Positive}
\acro{tn}[TN]{True Negative}
\acro{thor}[thor]{The House Of inteRactions}
\acro{tsdf}[TSDF]{truncated signed distance function}
\acro{vr}[VR]{Virtual Reality}
\acro{ycb}[YCB]{Yale-CMU-Berkeley}

\acro{ar}[AR]{augmented reality}
\acro{ate}[ATE]{absolute trajectory error}
\acro{bvip}[BVIP]{blind or visually impaired people}
\acro{cnn}[CNN]{convolutional neural network}
\acro{c2f}[c2f]{coarse-to-fine}
\acro{fov}[FoV]{field of view}
\acro{gan}[GAN]{generative adversarial network}
\acro{gcn}[GCN]{graph convolutional Network}
\acro{gnn}[GNN]{Graph Neural Network}
\acro{hmi}[HMI]{Human-Machine-Interaction}
\acro{hmd}[HMD]{head-mounted display}
\acro{mr}[MR]{mixed reality}
\acro{iot}[IoT]{internet of things}
\acro{llff}[LLFF]{Local Light Field Fusion}
\acro{bleff}[BLEFF]{Blender Forward Facing}

\acro{lpips}[LPIPS]{learned perceptual image patch similarity}
\acro{nerf}[NeRF]{neural radiance fields}
\acro{nvs}[NVS]{novel view synthesis}
\acro{mlp}[MLP]{multilayer perceptron}
\acro{mrs}[MRS]{Mixed Region Sampling}

\acro{or}[OR]{Operating Room}
\acro{pbr}[PBR]{physically based rendering}
\acro{psnr}[PSNR]{peak signal-to-noise ratio}
\acro{pnp}[PnP]{Perspective-n-Point}
%
\acro{sus}[SUS]{system usability scale}
\acro{ssim}[SSIM]{similarity index measure}
\acro{sfm}[SfM]{structure from motion}
\acro{slam}[SLAM]{simultaneous localization and mapping}

\acro{tp}[TP]{True Positive}
\acro{tn}[TN]{True Negative}
\acro{thor}[thor]{The House Of inteRactions}
\acro{ueq}[UEQ]{User Experience Questionnaire}
\acro{vr}[VR]{virtual reality}
\acro{who}[WHO]{World Health Organization}
\acro{xr}[XR]{extended reality}
\acro{ycb}[YCB]{Yale-CMU-Berkeley}
\acro{yolo}[YOLO]{you only look once}
\end{acronym} 


\firstsection{Introduction}

\maketitle

Allowing people to explore virtual replicas of physical environments has captivated interest for years. There are countless interesting places in the world worth capturing and exploring. Either to experience them from afar, to archive and document them, or to use them in applications for education or even games. However, high-quality experiences usually require talented 3D artists or expensive equipment such as laser scanners. Recent advances in generative models, including neural rendering and radiance fields, enable the creation of 3D worlds from photos alone e.g., \ac{nerf}~\cite{mildenhall_nerf_2020}, \ac{ngp}~\cite{muller_instant_2022}, or \ac{gs}~\cite{kerbl_3d_2023}. These approaches can be used to create high-quality representations of an object or even a full 3D scene. \Ac{gs} especially reduces rendering time~\cite{kerbl_3d_2023}, making it particularly suitable to use in \ac{vr}. 

By integrating \ac{gs} in \ac{vr}, users can experience nearly photorealistic environments. \Ac{nvs} enables the generation of renderings from novel viewpoints without the need to directly capture that specific part of the scene. This is particularly of interest for \ac{vr}, where users want to move outside the originally captured camera path. As \ac{gs} builds on primitives (splats), also used in traditional rendering, they can be seamlessly integrated with modelled objects such as 3D assets provided by game engines. This integration combines the strengths of \ac{gs} and game engines, allowing parts of the scene to be enhanced or made more interactive by replacing them with game engine content. 

For almost any editing of Gaussians and integration of 3D assets, the Gaussians must be separated into different classes so that they can be individually edited, removed, or replaced. 

Current approaches separating Gaussians primarily focus on ``circling" or forward-facing scenes~\cite{kerr_lerf_2023,ye_gaussian_2023,silva_contrastive_2024,mildenhall_local_2019}. Existing datasets concentrate on \ac{nvs} evaluation rather than offering pleasing \ac{vr} experiences. Applying \ac{nvs} to non-``circling" scenes introduces unique challenges for \ac{gs} segmentation. Non-``circling" scenes enable users to be surrounded by the 3D environment instead of viewing an isolated reconstruction. \ac{gs} on scenes with individual objects captured in a circular camera motion can use conventional classifiers and a convex hull to extract objects~\cite{ye_gaussian_2023}. These extracted scene parts can be used in \ac{vr}~\cite{xu_vr-nerf_2023,ye_gaussian_2023}. When applying \ac{gs} to scenes captured in a forward motion a simple convex hull or similar envelope-based segmentations cannot be used to remove objects, as the removal of an object may contain additional neighboring objects. Scenes recorded in forward motion only provide selected views (e.g. the front view) of parts of the scene. If object boundaries are not clear in this view, a removal over classifiers and envelopes tends to contain neighboring objects. For outdoor scenes, the similarity of features in the outdoor environment (e.g., reflective water) makes segmentation increasingly more difficult than segmentation of human-created scenes.

In this paper, we propose a novel Semantics-Controlled \ac{gs} approach (\approach) that enables precise segmentation of scene elements. This precise segmentation allows editing the scene by removing or replacing objects with other 3D assets. Examples include replacing large scene parts, like static reconstructed water or skies with matching (dynamic) 3D assets, facilitating a more customized experience, see \autoref{fig:teaser}. Allowing for the replacement of the sky, our approach can target inconsistent or unwanted weather conditions that may occur using pre-captured images. Additionally, replacing a cloudy sky with a clear blue sky allows for a more appealing \ac{vr} experience. 

We demonstrate and evaluate our novel approach using challenging non-``circling'' outdoor datasets. Examples for the various challenges posed to \ac{nvs} are small leaves or reflections in the water. Specifically, we provide a technical evaluation, showing that our approach outperforms the state-of-the-art in 3D separable \ac{gs}. Our segmentation performance is on par with other semantic \ac{nvs} approaches on the established 3D-OVS dataset. We also explore the advantages and disadvantages of combining our large-scale 3D asset generation technique with 3D assets from a Game Engine, where a significant and consistently dynamic element is replaced by an asset from the Game Engine. With respect to user experience, we compared video-based scene experience, plain \ac{gs}, and \approach bound to the camera capture path in an exploratory study. In our main study, we then compared plain \ac{gs} and \approach. Therein, the user was allowed to move freely and was thus able to take a closer look at the environment.

Our work makes the following contributions:
\begin{itemize}
\item A state-of-the-art approach for Semantic-Controlled Gaussian Splatting, namely (\approach), surpassing existing work.
\item A publicly available and challenging outdoor \ac{nvs} dataset with semantic labels\footnote{Dataset: \url{https://osf.io/s9uvy/?view_only=eff198d8752840e69a9f2b8c1c10b0a0}.}.
\item A comprehensive technical evaluation of our approach. 
\item A user study evaluating user experience and the users' perception on \approach.
\end{itemize}

Overall, our work is of specific relevance when using \ac{gs} to generate large-scale virtual environments beyond single objects, such as 3D reconstructions, that can be utilized for cultural and environmental purposes. The scope of application ranges from historical sites or regions threatened by climate change to exploring \ac{vr} as a sustainable alternative to physical tourism, enabling users to explore destinations from the comfort of their own space. Moreover, our work generally contributes to the rapidly progressing improvements of \ac{gs} with potential applications extending beyond content generation for \ac{vr} such as films, or games. 

\section{Related Work}

Experiencing, creating, and exploring a virtual space can either be done classically, using video replay~\cite{baker_casualstereo_2020}, panorama images~\cite{waidhofer_panosynthvr_2022,tong_applying_2024} and single-image-based depth enhancement~\cite{ajisa_3d_2024,pintore_panoverse_2023}, or in 3D using classic 3D reconstruction~\cite{dickson_vrvideos_2022,schieber_modular_2023} or radiance fields~\cite{kerbl_3d_2023,mildenhall_nerf_2020}.

\subsection{Virtual Reality Scene Content}

A common method for exploring static \ac{vr} content are panoramas~\cite{tong_applying_2024,baker_casualstereo_2020,zollmann_casualvrvideos_2020,waidhofer_panosynthvr_2022}. However, plain panoramas lack immersion as they miss depth information~\cite{bertel_omniphotos_2020,bertel_depth_2020}. Bertel et al.~\cite{bertel_omniphotos_2020} optimize this using 3D proxy fitting. 
 Ajisa et al.~\cite{ajisa_3d_2024} propose inpainting to view an indoor or outdoor scene based on a single panorama image, thus limiting the area of movement to one area of a scene.

Other approaches enriching outdoor photographs~\cite{snavely_photo_2006,martin-brualla_nerf_2021,boming_zhao_and_bangbang_yang_factorized_2022,freer_novel-view_2022} do not directly address \ac{vr}. Freer et al.~\cite{freer_novel-view_2022} separate people in front of sightseeing attractions and utilize neural rendering to inpaint the area. Zhao et al.~\cite{boming_zhao_and_bangbang_yang_factorized_2022} integrate the capture of one person in sight and extrapolate it using online data.

Apart from simply replaying a scene, advancements in deep learning allow the generation of neural content for \ac{vr}. 

Campos et al.~\cite{campos_procedural_2019} utilize procedural content generation based on agents and decision trees to enable a unique user experience for each user. \Acp{llm}~\cite{aguina-kang_open-universe_2024,yang_llm-grounder_2024,yin_text2vrscene_2024} and other foundation models~\cite{chen_meshxl_2024} further improved content generation and can be utilized to create content ranging from simple text to 3D~\cite{yang_llm-grounder_2024,yin_text2vrscene_2024}. While standard \ac{llm}s are challenged to create \ac{vr} scenes,  Yin et al.~\cite{yin_text2vrscene_2024} propose Text2VRScene, generating synthetic non-photorealistic, but content aware \ac{vr} scenes.

\subsection{Novel View Synthesis}

Classically, radiance field-based approaches do not directly target large-scale scene extraction for \ac{vr}. Scenes created with radiance fields can be used for virtual content~\cite{deng_fov-nerf_2022,li_rt-nerf_2022,jiang_vr-gs_2024,rolff_interactive_2023} and multiple \ac{mr} devices allow to generate such virtual  content~\cite{schieber_nerftrinsic_2023}.
However, to use radiance field-based rendering in \ac{vr}/\ac{xr} challenges include scene representation and the underlying data structure. For example, changing 3D scene content may prove difficult as editing a \ac{nerf} is not trivial~\cite{chen_omnire_2024,wang_clip-nerf_2022,yang_learning_2021}. ClipNeRF~\cite{wang_clip-nerf_2022} addresses this by adapting an existing \ac{nerf} in a separate training step. While \ac{nerf} is advantageous for \ac{nvs}, its real-time rendering capacity for \ac{vr} has been outperformed by \ac{gs}~\cite{kerbl_3d_2023}.

\begin{figure*}[t!]
    \centering
    \includegraphics[width=0.9\textwidth]{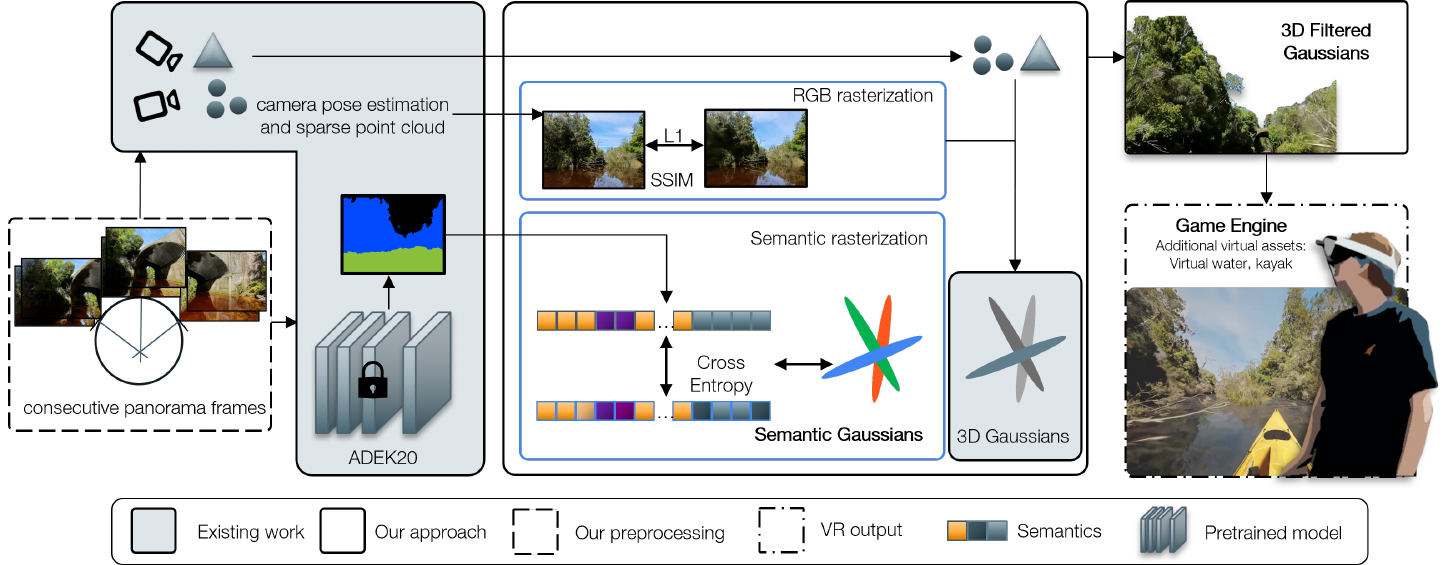}
    \caption{\textbf{Architecture of \approach.} We first extract images from our continuous panoramic stream. Using COLMAP to estimate the camera positions, we obtain the sparse point cloud for the initialization of \ac{gs}. To enable 3D filtering, the data is preprocessed with a segmentation model. During 3D Gaussian training, we use \acs{ce}-loss, L1-loss and \acs{ssim}-loss to fit our scene into the RGB and segmentation space. The final 3D representation can be viewed in the viewer, or individual parts of the scene be extracted and used in \ac{vr}.}
    \label{fig:arch}
\end{figure*}

Moreover, \ac{gs} is an explicit representation of the scene, allowing its easier adaptation compared to implicit \ac{nerf} representation. A \ac{gs} scene starts with a sparse point cloud. Using photometric loss as well as densifying and pruning steps, the scene is refined. The initial \ac{gs} representation can be challenged by large \ac{slam} like scenes~\cite{kerbl_3d_2023,kerbl_hierarchical_2024}. To overcome this, Kerbl et al.~\cite{kerbl_hierarchical_2024} introduce hierarchical \ac{gs}, enabling a block-wise optimization for larger scenes depending on the camera location at rendering time.

For \ac{vr} Jian et al. propose VR-GS~\cite{jiang_vr-gs_2024} for indoor scenes. VR-GS integrates inpainting in the \ac{gs} training process, using  mesh exports and manual post-processing for each object. Thus, the objects are movable in \ac{vr}. Chen et al.~\cite{chen_omnire_2024} reconstruct dynamic urban scenes using Gaussian scene graphs. Each graph holds information about individual parts of the scene.

Apart from separating a scene based on statics and dynamics, another line of work is semantic \ac{gs}. Semantic \ac{gs} enables extracting parts of a scene as 3D assets. Feature 3DGS~\cite{zhou_feature_2023} utilizes \ac{sam} embeddings to improve \ac{nvs} quality. Similarly, Gaussian Grouping~\cite{ye_gaussian_2023} introduces semantic features into a \ac{gs} structure, proposing identity encoding allowing to group 3D Gaussians. Building on \ac{sam}-DEVA~\cite{cheng_tracking_2023},  Contrastive Gaussian Grouping~\cite{silva_contrastive_2024} extends this idea by omitting the tracking step in pre-processing and identifying consistence labels through a contrastive learning step. 
Disadvantageously, when interacting in 3D and especially when exporting the scene content to \ac{vr}, not all formats or output types can be used. 

\subsection{User Experience of Reconstructed Environments}

\ac{nvs} has been explored in \ac{xr}, specifically for \ac{mr}~\cite{mohr_mixed_2020, sakashita_sharednerf_2024} with screen-based applications, and in \ac{vr} with individual 3D reconstructed parts of a scene~\cite{kleinbeck_neural_2024}.

Sakashita et al.~\cite{sakashita_sharednerf_2024} visualize a point cloud and a \ac{nerf} using a head-mounted camera and a desktop computer for shared interactions. The desktop computer visualizes the point cloud overlaying the \ac{nerf}. In a preliminary user study they detected a preference for \acp{nerf} combined with point cloud overlay in comparison to video or pure point cloud visualization.

The use of 3D assets for task execution planning benefits from 3D assets generated with \ac{sdf} based approaches~\cite{kleinbeck_neural_2024,li_neuralangelo_2023}. 
Kleinbeck et al.~\cite{kleinbeck_neural_2024} create a digital twin of operating rooms in \ac{vr}. Using \ac{sdf}-based mesh reconstruction of the scene, an accurate mesh is created. By manually post-processing individual scene parts, a \ac{vr} experience is created that can be explored by participants.

\subsection{Research Gap}

Although recreating the real-world using a video or photo~\cite{ajisa_3d_2024,tong_applying_2024,teo_mixed_2019} achieves the most realism, 3D reconstruction enables more freedom in \ac{vr} by allowing the user to move outside of the captured camera trajectory. What is needed is an approach that covers both the processing of \ac{gs} for \ac{vr} and a dataset that allows both a pleasant virtual experience and the processing of \ac{nvs} and scenes. Existing semantic \ac{gs} approaches normally focus on a scene, where a camera ``circles'' around one bigger object~\cite{knapitsch_tanks_2017} or multiple smaller objects~\cite{ye_gaussian_2023,silva_contrastive_2024,qin_langsplat_2023}, as well as feature-based separation~\cite{ye_gaussian_2023,silva_contrastive_2024}. These approaches often concentrate on data with clear boundaries between objects due to lower feature similarity.

With \approach and our \ac{nvs} dataset we directly address this research gap, enabling accurate scene editing and extraction of large scene parts. Moreover, our dataset captures pleasing outdoor scenes in a non-``circling'' setup, enabling \ac{vr} experiences where the user is surrounded by the virtual environment. 

\section{Method}

Our approach, \approach, separates Gaussians into segmentation classes, directly assigning the respective segmentation class. To achieve this, we alter the Gaussian rasterization process. This allows the classification of 3D Gaussians in the 2D image space and 3D Gaussian space at almost equal quality, which is advantageous in non-``circling'' setups. The direct class assignment of \approach enables the removal of complete classes at a large-scale, while omitting feature similarity\footnote{Dataset: \url{https://osf.io/s9uvy/?view_only=eff198d8752840e69a9f2b8c1c10b0a0}.}.

\subsection{Semantics-Controlled Gaussian Splatting}

\subsubsection{Preliminary 3D Gaussians}
3D \ac{gs}~\cite{kerbl_3d_2023} represents an explicit scene representation initialized from a (sparse) point cloud. For the Gaussian representation $\Sigma'$ represents the 2D rasterized Gaussians, $J$ is the Jacobian of the affine approximation,  $W$ is the world-to-camera transformation matrix and $\Sigma$ is the 3D representation.

\begin{equation}
\Sigma' = J W \Sigma W^T J^T
\end{equation}

Each Gaussian $G$ is represented by its 3D center position $(x)$ and a 3D covariance matrix $(\Sigma)$ that can be denoted as a rotation matrix and scaling matrix. To represent colors and scene appearance, each Gaussian holds a density value ($\sigma$)  and \ac{sh} coefficient to encode RGB information. To retrieve the color ($c$) of each pixel, alpha ($\alpha$) blending is used.

\begin{equation}
\textit{RGB} = \sum_{i \in \textbf{N}} T_i \alpha_i c_i \textit{~with~} T_i = \prod_{j=1}^{i-1}(1-\alpha)
\end{equation}

\begin{figure*}[t!]
    \centering
    \includegraphics[width=\textwidth]{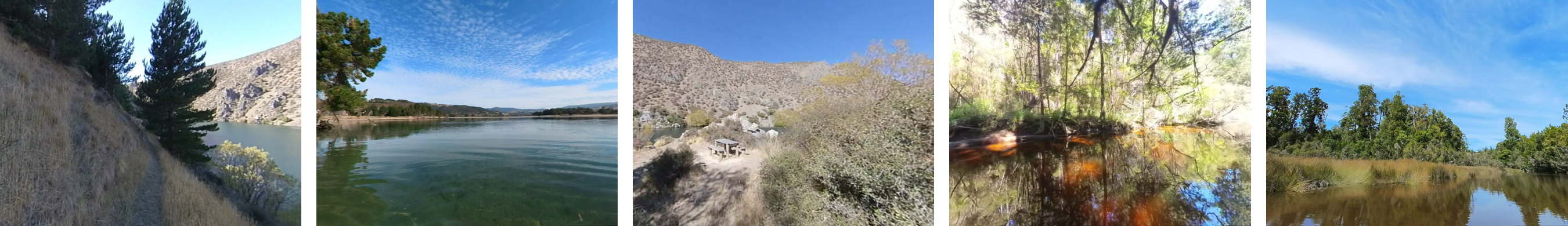}
    \caption{\textbf{Images of our dataset.} Tree, Open Sea, Picnic, Outback, and Kayak (from left to right).}
    \label{fig:dataset}
\end{figure*}

\begin{table*}[t!]
\centering
\caption{\textbf{Technical evaluation.} We report \acs{psnr}, \acs{ssim} and \acs{lpips}. The best results are highlighted in bold. Best results within a range of $\pm$ 0.5 dB are highlighted in \colorbox[HTML]{D9EAD3}{light-green} and above 0.5 improvement in \colorbox[HTML]{6AA84F}{dark-green}. Results worse than 1.0 compared to our approach are highlighted in \colorbox[HTML]{FFCC69}{orange}.}\label{tab:rendering_result_our_data}
    \begin{tabular}{l|ccc|ccc|ccc}
    \toprule
     \multirow{2}{*}{Approach} & \multicolumn{3}{c|}{\textbf{Gaussian Grouping}~\cite{ye_gaussian_2023}}    & \multicolumn{3}{c|}{\textbf{Gaussian Grouping}}      &    \multicolumn{3}{c}{ \multirow{2}{*}{\textbf{OURS}}}  \\ 
     & \multicolumn{3}{c|}{\textbf{SAM DEVA (original)}}    & \multicolumn{3}{c|}{\textbf{OUR labels}}        \\ 
           & {PSNR} & {SSIM} & {LPIPS} & {PSNR} & {SSIM} & {LPIPS} & {PSNR} & {SSIM} & {LPIPS} \\ \midrule
        Tunnel  & \cellcolor[HTML]{FFCC69}{20.54} & 0.654 & 0.379 & \cellcolor[HTML]{D9EAD3}{\textbf{23.29}} & \textbf{0.792} & 0.403  & \cellcolor[HTML]{D9EAD3}{23.11} & 0.727 & \textbf{0.316} \\
        Lake & \cellcolor[HTML]{FFCC69}{20.80} & 0.703 & 0.329  & \cellcolor[HTML]{D9EAD3}{21.59} & 0.734 & 0.306  & \cellcolor[HTML]{D9EAD3}{\textbf{21.71}} & \textbf{0.735} & \textbf{0.303}  \\ 
        Kayak  & \cellcolor[HTML]{FFCC69}18.61 & 0.576 & 0.448  & \cellcolor[HTML]{FFCC69}21.51  & 0.662 & \textbf{0.406} & \cellcolor[HTML]{6AA84F}\textbf{22.13} & \textbf{0.667} & 0.415  \\
        Open Sea & \cellcolor[HTML]{FFCC69}27.82 & 0.837 & 0.352  & \cellcolor[HTML]{FFCC69}27.81  & 0.831 & 0.358  & \cellcolor[HTML]{6AA84F}\textbf{28.85} & \textbf{0.840} & \textbf{0.300}  \\
        Short Ride & \cellcolor[HTML]{FFCC69}18.67 & 0.635 & 0.374  & \cellcolor[HTML]{FFCC69}19.51 & 0.678 & 0.336  & \cellcolor[HTML]{6AA84F}\textbf{20.02} & \textbf{0.694} & 0.324  \\
        Outback & \cellcolor[HTML]{FFCC69}21.18 & 0.700 & 0.408  & \cellcolor[HTML]{FFCC69}24.10 & 0.763 & 0.335  & \cellcolor[HTML]{6AA84F}\textbf{25.13} & \textbf{0.799} & 0.299  \\ 
        Picnic    & \cellcolor[HTML]{FFCC69}23.96 & 0.795 & 0.241  & \cellcolor[HTML]{D9EAD3}24.90 & \textbf{0.811} & 0.225  & \cellcolor[HTML]{D9EAD3}\textbf{24.97} & 0.805 & 0.215  \\
        Tree   & \cellcolor[HTML]{FFCC69}23.29 & 0.792  & 0.403 & \cellcolor[HTML]{D9EAD3}25.40 & \textbf{0.814} &   0.358  & \cellcolor[HTML]{D9EAD3}\textbf{25.83}      & 0.802 & 0.357  \\ \midrule
        Mean   & \cellcolor[HTML]{FFCC69}21.86 & 0.712  & 0.367   & \cellcolor[HTML]{D9EAD3}23.51 & 0.761 & 0.34  & \cellcolor[HTML]{D9EAD3}\textbf{23.97} & 0.759 & \textbf{0.32}  \\ \bottomrule
    \end{tabular}
\end{table*}

\subsubsection{3D Gaussian Segmentation}

We enhance the \ac{gs} representation by integrating semantic information, extending the conventional RGB rasterization process to support semantic rendering, see \autoref{fig:arch}. This adaptation allows us to reformulate the segmentation problem within the Gaussian parameter space, facilitating the direct assignment of class IDs to each Gaussian in 3D space.

In our approach, the differentiable rendering pipeline first converts spherical harmonics (\ac{sh}) into RGB values, which are then splatted onto the 2D image plane. For the semantic map ($s$), a parallel rasterization process is used, where the \ac{sh} components are set to zero, effectively isolating the semantic attributes from  RGB.

This method diverges from traditional classification-based approaches~\cite{ye_gaussian_2023,zhou_feature_2023}, which typically require a separate classifier for semantic segmentation. Instead, our approach assigns class IDs consistently during training, using cross-entropy loss, addressing challenges related to feature space similarities, especially in outdoor scenes with significant reflections. The 3D segmentation is then projected onto a 2D map using alpha blending ($\alpha$):

\begin{equation}
\textsc{Segmentation} = \sum_{c \in C} T_i \alpha_i s_{c_i}
\end{equation}

Our approach modifies the rasterization process to facilitate backpropagation of the segmentation map, similarly to how RGB values are handled. After the rasterization step, each Gaussian is associated with a class segmentation ID that is splatted onto the 2D image plane. This enables the application of \ac{ce} loss to supervise the 3D \ac{gs} segmentation through a 2D loss function.

The loss function for semantic segmentation is defined as

\begin{equation}
{L}_{\textit{CE}} = -\sum{i} \sum_{c} s_{ic} \log \hat{s}_{ic}
\end{equation}

where \(s_{ic}\) is the ground truth segmentation for class \(c\) at pixel \(i\), and \(\hat{s}_{ic}\) is the predicted probability for class \(c\) at pixel \(i\).

Additionally, the assigned class ID allows for the selective removal of one or more sets of 3D Gaussians at a large-scale, enabling targeted modifications of the 3D scene.

\subsubsection{3D Gaussian Separation}

Our 3D Gaussian separation utilizes our direct segmentation ID class assignment to remove 3D Gaussians from the complete set of 3D Gaussians ($\mathcal{G}_{\textit{new}}$). Given the desired object class or object classes to remove ($c_{r=1..n}$), our approach enables removing one or more classes per scene. Since each Gaussian class has a direct identifier, no additional post-processing, as e.g. creating a convex hull~\cite{ye_gaussian_2023}, is required. Moreover our approach allows to remove directly large-scale object, see \autoref{fig:arch}.

The assigned segmentation class ID also allows for the selective removal or modification of one or more sets of 3D Gaussians.

\begin{equation}
\mathcal{G}_{\textit{new}}(x; \Sigma) = \mathcal{G}(x; \Sigma) \cdot \textbf{I}(s_{ic} \neq {c_{r=1..n}})
\end{equation}

where \(\textbf{I}\) is the indicator function, ensuring only the Gaussians not belonging to the removed class ID are retained.

\begin{figure*}[t!]
    \centering
    \includegraphics[width=0.9\textwidth]{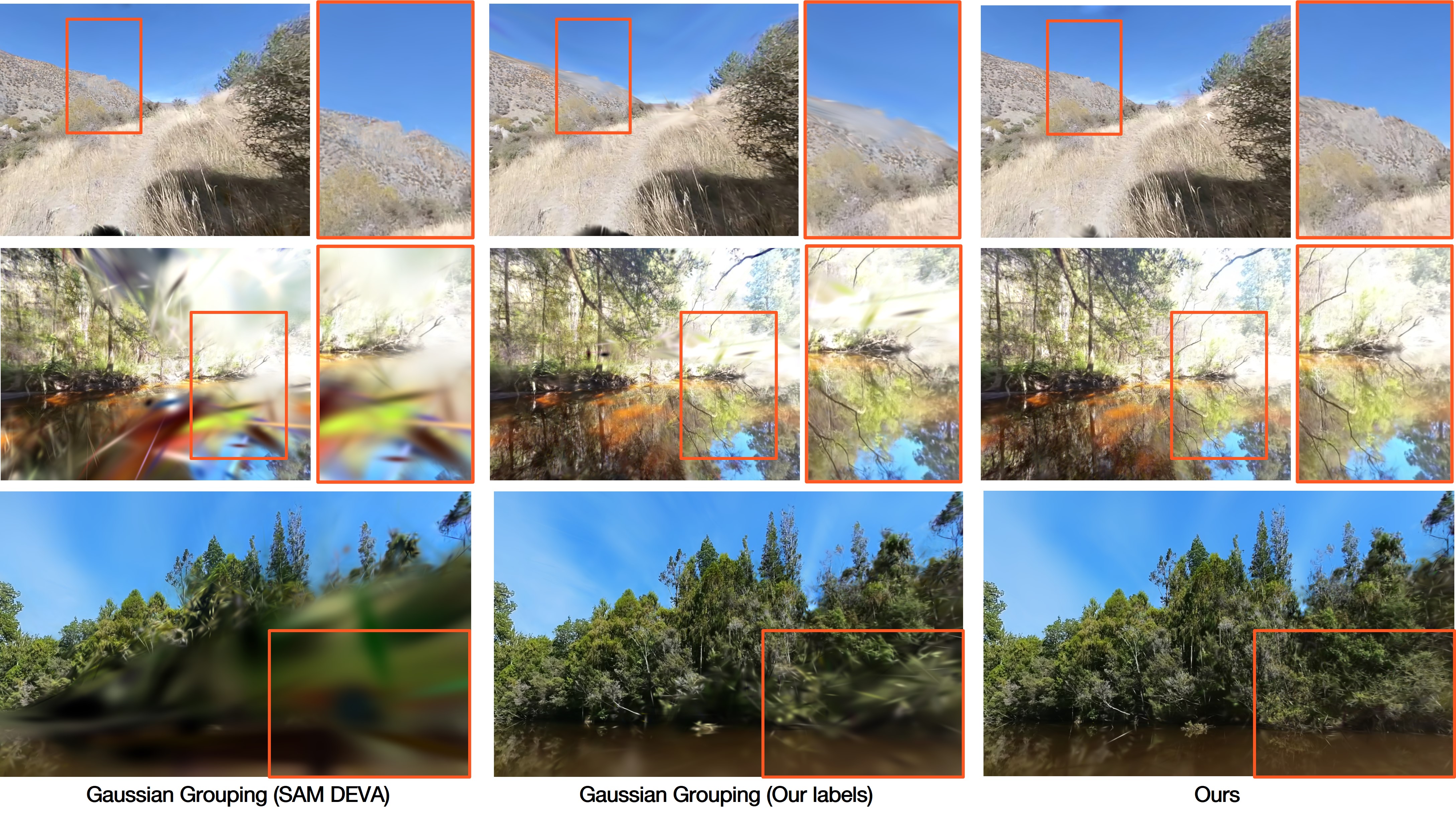}
    \caption{\textbf{Example comparison of Gaussian Grouping, Gaussian Grouping (improved labels) and our approach.} All scenes contain water, sky and vegetation. The hiking sequence (top), shows outliers on the mountain (transparency), the kayak outback scene (center) shows the challenges of the water and the kayak scene (bottom) shows the challenges of the closed stacked trees.}
    \label{fig:pure}
\end{figure*}

\subsection{Large-Scale Outdoor 3D Asset  Dataset}

Existing semantic \ac{nvs} datasets focus on indoor scenes~\cite{liu_weakly_2023,ye_gaussian_2023,kerr_lerf_2023} following a circling camera path. We propose a dataset, which provides challenging outdoor scenes containing reflective surfaces, similar features and challenging structures (trees, leaves, water). Our dataset is captured using Insta360 cameras of types X1, X2 and X3. The camera is positioned in front of individuals engaged in various activities, like kayaking. Employing a panoramic setup, we derive multiple camera poses from the resulting forward moving video stream. Example images can be seen in \autoref{fig:dataset}. By combining forward-facing images with those angled $\pm60$/$\pm30^{\circ}$ to the left and right and $\pm 10^{\circ}$ up and down, we achieve comprehensive coverage of the scene. For privacy reasons this setup excludes the individual experiencing the activity. After extracting images from the video stream we retrieve segmentation masks using DPT~\cite{ranftl_vision_2021}. Our outdoor recordings feature known classes. Therefore, we use the ADEK20 labels. Afterwards, the camera poses of the image set are retrieved~\cite{schonberger_structure--motion_2016}. 

We split our dataset into two categories, pure \ac{nvs}, with images angled  $\pm60^{\circ}$/$\pm30^{\circ}$, and another set in which we provide the full $360^{\circ}$ video to enable comparisons of classic $360^\circ$ videos and \ac{nvs} in \ac{vr}. For the images angled $360^\circ$ we create a stacked video using $360^\circ$ monodepth~\cite{rey-area_360monodepth_2022,ajisa_3d_2024}.

\section{Technical Evaluation}

Our approach groups individual Gaussians based on their directly regressed semantic class. Since our approach aims to separate the 3D Gaussian's, we compare it on our dataset with identity encoding, namely Gaussian-Grouping~\cite{ye_gaussian_2023}. On the 3D-OVS dataset, we assess segmentation quality and compare it with other state-of-the-art novel view segmentation approaches building upon language supervision~\cite{qin_langsplat_2023,kerr_lerf_2023} and contrastive learning~\cite{silva_contrastive_2024}.

\subsection{Metrics}

To compare the rendering quality of the novel views, we report \ac{psnr}, \ac{ssim}~\cite{wang_image_2004} and \ac{lpips}~\cite{zhang_unreasonable_2018}. For the segmentation performance, we report \ac{miou}.

\subsection{Implementation Details}

We used ffmpeg to extract images from the video stream. For camera pose retrieval and sparse reconstruction we leverage COLMAP~\cite{schonberger_structure--motion_2016}.

Our approach is implemented in Python using PyTorch and CUDA. All scenes of our dataset can be trained on one single RTX4090 with 24GB VRAM using our approach. The comparing methods were trained on an A100 with 40GB VRAM as they required more VRAM.

\subsection{Novel View Synthesis Quality}

Our approach improves \ac{nvs} quality on outdoor scenes, see \autoref{fig:pure} and \autoref{tab:rendering_result_our_data}. In our scenario, continuous labels and classes are available a priori, allowing us to conduct a direct and fair comparison with a classifier-based method~\cite{ye_gaussian_2023}. As highlighted in \autoref{tab:rendering_result_our_data}, we outperform the baseline using continuous labels from \ac{sam} DEVA~\cite{cheng_tracking_2023} on all scenes. When comparing the segmentation maps used by Gaussian Grouping and our segmentation maps, a clear difference in quality is visible. Our preprocessing for outdoor semantic segmentation produces a better quality.
 
Consequently, for a fairer comparison, we updated the segmentation maps from Gaussian Grouping with our segmentation maps. We retrained Gaussian Grouping using our enhanced labels. As denoted in \autoref{tab:rendering_result_our_data}, the improved labels strongly enhance the \ac{nvs} performance of Gaussian Grouping. This can be seen in \autoref{fig:pure}. Nevertheless, our approach outperforms both standard Gaussian Grouping and Gaussian Grouping using our improved labels on seven out of eight scenes on our outdoor dataset, see \autoref{tab:rendering_result_our_data}. Moreover, we outperform it on five scenes in \ac{ssim} and seven scenes in \ac{lpips}. The visual improvement is also visible when comparing the images in \autoref{fig:pure}.

\begin{figure*}[t!]
    \centering
    \includegraphics[width=0.9\textwidth]{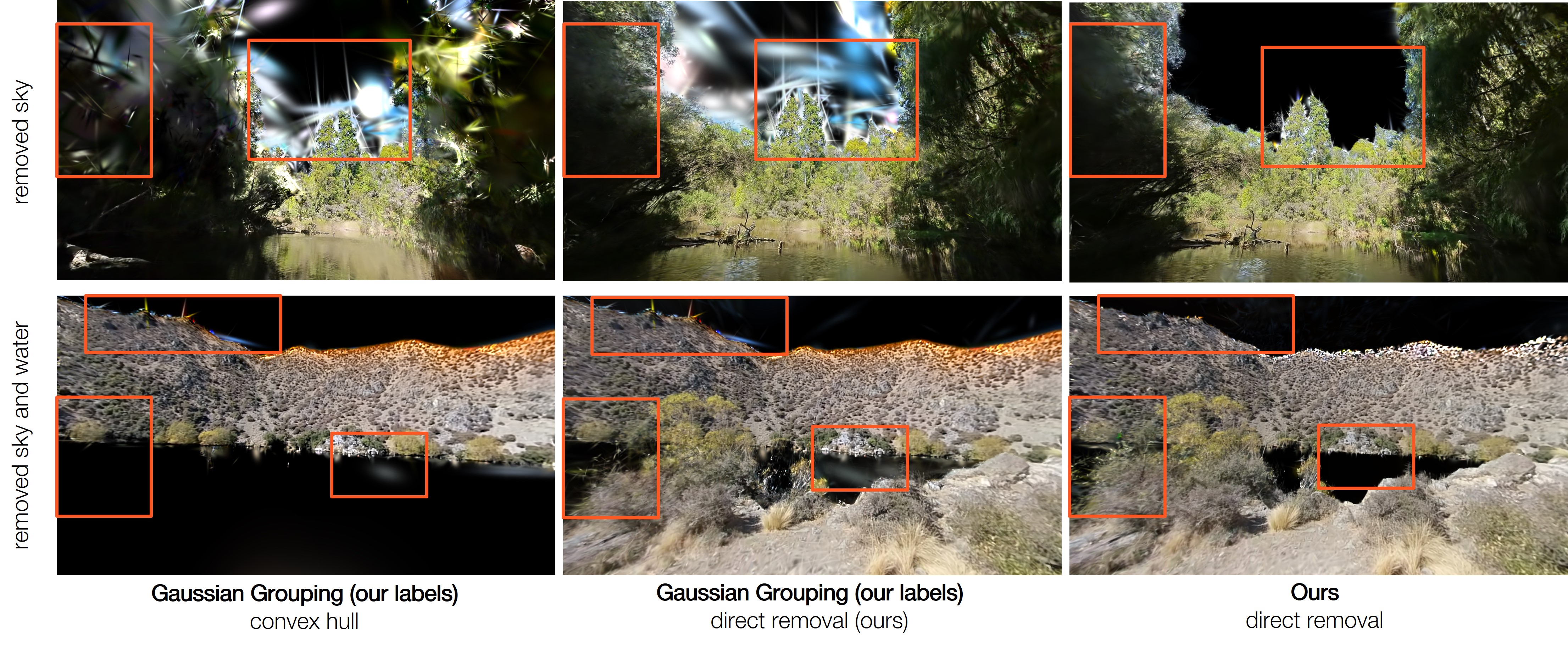}
    \caption{\textbf{Class removal on our dataset.} The convex hull removes to much of the scene (left), using the same direct removal (center) as in our case, leads to more outliers, compared to our approach (right).}
    \label{fig:removal}
\end{figure*}

\subsection{Segmentation Performance}

We distinguish the segmentation performance into segmentation influencing object removal and into classic segmentation performance in terms of \ac{miou}.

\subsubsection{Large-Scale Object Removal}

The benefit of \approach becomes even clearer through the use of post-processing steps to remove Gaussians. As shown in \autoref{fig:removal}, \approach evidently better removes the Gaussians compared to the baseline.  \ac{sam} DEVA is challenged by the outdoor scenario and the inconsistent labels lead to a degradation in \ac{nvs} quality. However, even with improved labels increasing the \ac{nvs} quality, the achieved performance in object removal is not on par with our approach, see \autoref{fig:removal}. We even tested our object removal in \autoref{fig:removal} for the baseline. Still, our approach shows a better result.

\approach can remove individual classes and shows noticeably clearer and better boundaries to other objects/classes in the scene. This leads to a higher-quality scene which can be integrated into Game Engines. As can be seen in the top line in \autoref{fig:removal}, the sky and the tree are too connected when using a convex hull. Since we do not use a classic circular capturing setup here, a convex hull may not be the best way to remove unwanted objects. Therefore, we propose direct removal by class. As the comparison in \autoref{fig:removal} shows, our approach better distinguishes the individual classes and removes large-area parts directly and accurately.

\begin{table}[t!]
    \centering
        \caption{\textbf{Evaluation on the 3D-OVS dataset~\cite{liu_weakly_2023}.} We report \ac{miou} per class and overall.}
    \resizebox{\columnwidth}{!}{
    \begin{tabular}{l|ccccc|c} \toprule
       \textbf{Approach}	&	Bed 	&	Bench 	&	Room 	&	Sofa 	&	Lawn 	&	\textbf{Mean}	\\ \midrule
    LERF~\cite{kerr_lerf_2023} 	&	73.5	&	53.2	&	46.6	&	27.0	&	73.7	&	54.8	\\
    Gaussian Grouping~\cite{ye_gaussian_2023} 	&	\textbf{97.3}	&	73.7	&	79.0	&	{68.1}	&	\textbf{96.5}	&	82.9	\\
    LangSplat~\cite{qin_langsplat_2023} 	&	34.3	&	84.8	&	56.3	&	67.7 &	95.8	&	67.8	\\
    Contrastive Grouping~\cite{silva_contrastive_2024} 	&	95.2	&	\textbf{96.1}	&	\textbf{86.8}	&	67.5	&	91.8	&	87.5	\\ \midrule
    Ours	&	94.4	&	89.8	&	73.2	&	\textbf{92.5}	&	89.0	&	\textbf{87.8}	\\ \bottomrule
    \end{tabular}}

    \label{tab:lerf_miou}
\end{table}

\subsubsection{Segmentation on 3D-OVS}

We compare \approach with Gaussian-Grouping~\cite{ye_gaussian_2023}, LERF~\cite{kerr_lerf_2023}, LangSplat~\cite{qin_langsplat_2023} and Contrastive Gaussian Grouping~\cite{silva_contrastive_2024} on the state-of-the-art 3D-OVS dataset~\cite{liu_weakly_2023}. We report \ac{miou} per scene and the \ac{miou} overall scenes in Table \ref{tab:lerf_miou}. As reported in Table \ref{tab:lerf_miou}, we outperform existing work on one out of five scenes and perform competitively on all other scenes. The improvement on the ``Sofa'' scene of the 3D-OVS shows that we outperform existing work in the overall \ac{miou}.

\begin{figure}[t!]
    \centering
    \includegraphics[width=\columnwidth]{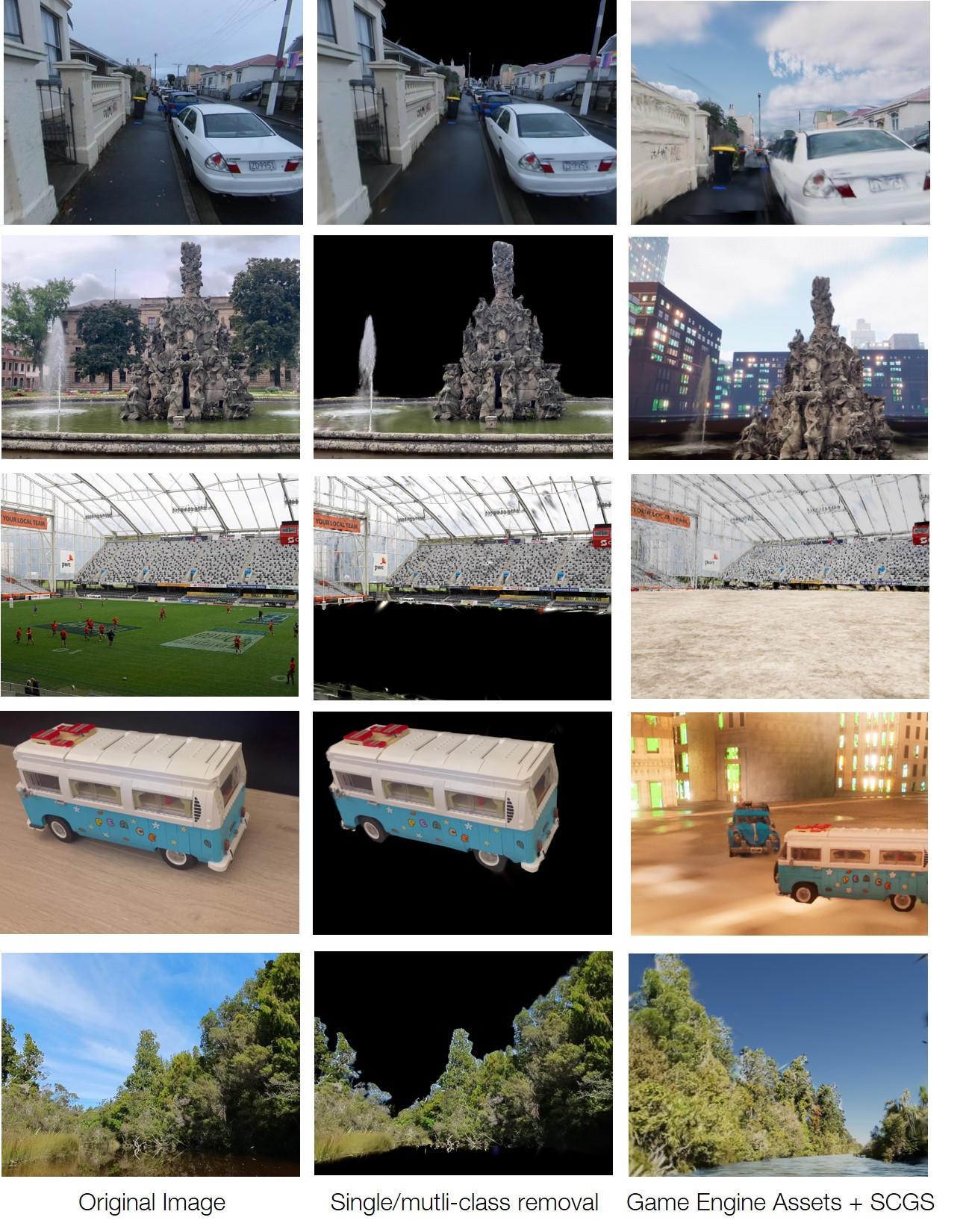}
    \caption{\textbf{Use Cases.} Our approach can be applied to various cases of large-scale scene removal/editing: Sky replacement (top row, second, bottom row), scenes outside of our dataset like sport fields or fountains (second row), or smaller objects like brick cars (fourth row). We display images from the original capturing (left), the removed class in black (center), and the Game Engine enriched scene (right) from a novel viewpoint. }
    \label{fig:usecasesexample}
\end{figure}

\subsection{Use Cases}

\approach has broad applicability in Game Engine environments, supporting a range of use cases. The primary objective, large-scale asset generation, addresses the needs of diverse virtual environments. This can be particularly valuable for games or virtual experiences in the fields like virtual tourism, where specific assets, such as nature, sports fields, or famous statues, need to be seamlessly integrated into virtual worlds. Additionally, advertisement signs can be replaced using \approach to avoid copyright issues. As shown in \autoref{fig:usecasesexample}, both single classes (e.g., sky) or multiple classes (e.g., sky and water, or sky and buildings) can selectively be removed. \approach enables the incorporation of new assets from Game Engines, allowing novel viewpoints and more dynamic scene rendering.

\section{User Study}

To investigate user perceptions of plain \ac{gs} and \approach (\approach combined with 3D assets) in \ac{vr}, we conducted an exploratory and a main user study using a within-subject (repeated measures) design. The ethical approval of the participating institutions was granted.

\subsection{Apparatus}

We used an Oculus Quest 3 \ac{hmd} connected via Oculus Link to a workstation powered by an NVIDIA RTX 4090. Rendering was done on the workstation in Unreal Engine using the Lumalab plugin~\cite{luma} for \ac{gs} and custom scene setups.

\subsection{Procedure}
\label{subsec:procedure}

After welcoming participants and obtaining consent, they completed a questionnaire on demographics and \ac{vr} experience. Followed by familiarizing them with the \ac{hmd}. Then they experienced the conditions in a randomized, balanced order, filling out a questionnaire after each one. At the end, they ranked the conditions. 

\subsection{Analysis Strategy}

All analyses of the user studies were performed using RStudio Version 4.4.1. We evaluated the study using one-way repeated measures ANOVA where suitable (three conditions), a paired samples t-test (two conditions), and Tukey's post-hoc analysis with Bonferroni correction where suitable. Our significance level is set to 0.05.

We applied the Shapiro-Wilk test to test for normal distribution.

\subsection{Explorative Study}

According to the literature~\cite{teo_mixed_2019}, $360^\circ$ panorama images/videos enhance users' sense of presence, but research on perceived realism and presence in \ac{gs} is limited. In our preliminary study, we focused on these aspects using $360^\circ$ RGB-D video as a baseline. 

As existing work on \ac{gs} provides image metrics or \ac{vr} examples without specific user feedback, the perceived presence in a \ac{gs} environment is so far unknown. The goal of this explorative study is to establish a frame of reference within which we will operate in our main study in which we give the user more freedom.

\paragraph{Conditions}

The original video was recorded in a seated kayak scenario, so we recreated this environment for our study by adding a virtual kayak to both the plain \ac{gs} and the \approach scene. Our baseline was a $360^\circ$ panorama video with generated depth (condition 1)~\cite{rey-area_360monodepth_2022, ajisa_3d_2024}. The other two conditions were plain \ac{gs} without dynamics (condition 2) and \approach with added water dynamics (condition 3). Throughout the experience, the user followed the camera path at the center of a river seated in a (virtual) kayak.

\paragraph{Measures}

To measure the perceived presence in \ac{vr}, we employed the \ac{ipq}~\cite{schubert_sense_2003,tran2024classifying}. To evaluate the preference of each participant, we asked our additional questions rating the environment, at the end we let the participant rate their preferred condition.

\paragraph{Participants}
We recruited 24 participants (14 male, 9 female, 1 non-binary) through announcements, notice boards and word-to-mouth. The participants had an average age of 22.42$\pm$4.93). 

\begin{figure}[t!]
    \centering
    \includegraphics[width=0.6\columnwidth]{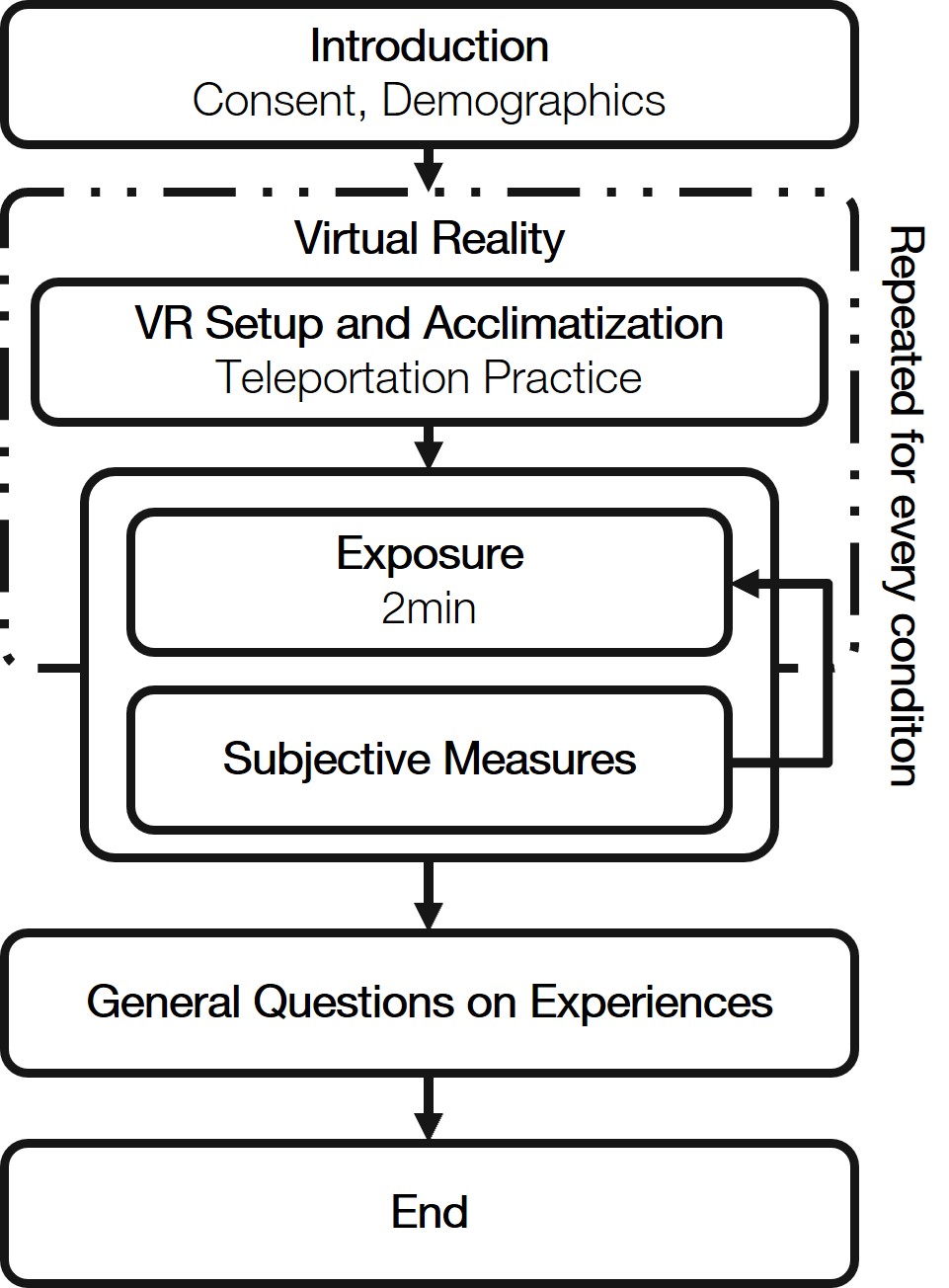}
    \caption{\textbf{User Study Procedure Diagram.}}
    \label{fig:user_study}
\end{figure}

\paragraph{Results and Discussion}

We found a significant difference in ``realism" when comparing the video condition with plain \ac{gs} and the video with our approach (\approach) see \autoref{tab:kayak_pilot_ipq}. Applying Tukey's post-hoc analysis and pairwise t-tests with Bonferroni correction, we reveal a significant difference between video compared to \ac{gs} ($p<0.028$) and video compared to our approach ($p<0.016$). 

\approach ranked second for first preference and highest for second preference. The video condition scored first rank.
The evaluation revealed that plain \ac{gs} scores the lowest in terms of user preference. 

We found a significant difference between the video condition and both \ac{gs} and \approach. This is reasonable as the video, where users follow the original camera path, naturally looks most realistic in terms of image quality. \approach performed similarly to plain \ac{gs}, which is supported by similar median and standard deviation. Our approach ranked second in preference, after the video condition, while plain \ac{gs}, the second condition ranked last.

The \ac{ipq} does not reflect all feedback, as participants expressed a desire for free movement and described the $360^\circ$ video as flat and resembling 2D content. Plain \ac{gs} was criticized for lacking immersion. In contrast, comments like \textit{“The moving water in the river had a huge impact, it felt so realistic.”} suggest positive feedback for \approach, particularly regarding the added dynamic assets like flowing water and reflections. These findings point to the potential benefits of \approach, indicating a need for further exploration in our main study. In the main study, the participants could move freely instead of being seated in the virtual kayak.

\begin{table}[t!]
    \centering
     \caption{\textbf{Results of the \ac{ipq} for the explorative study.}}
    \resizebox{\columnwidth}{!}{
    \begin{tabular}{l|ccccccccc} \toprule
         \ac{ipq} & $M_{video}$ & $SD_{video}$ &  $M_{GS}$ & $SD_{GS}$ & $M_{\approach}$ & $SD_{\approach}$ & $F$ & $p$ \\ \midrule
         General Presence & 3.83 & 1.37 & 3.38 & 1.70 & 3.42 & 1.63 & 0.595 &  0.554\\
         Spatial Presence & 3.68 & 1.14  & 3.42 & 1.11 & 3.39 & 1.14 & 0.189 & 0.828 \\
         Involvement & 3.35 &  1.32  & 3.27 & 1.24 & 3.26 & 1.29 & 0.009 & 0.991\\
         Realism & 2.86 & 0.92 & 2.05 & 0.87 & 2.02 & 0.95 & 5.145 & \textbf{0.008}\\ \midrule
         Overall & 3.37 & 0.95 & 2.98 & 0.85 & 2.96 & 0.92 & 0.921 &  0.403\\ \bottomrule
    \end{tabular}}
    \label{tab:kayak_pilot_ipq}
\end{table}

\begin{figure*}[t!]
    \centering
    \includegraphics[width=\textwidth]{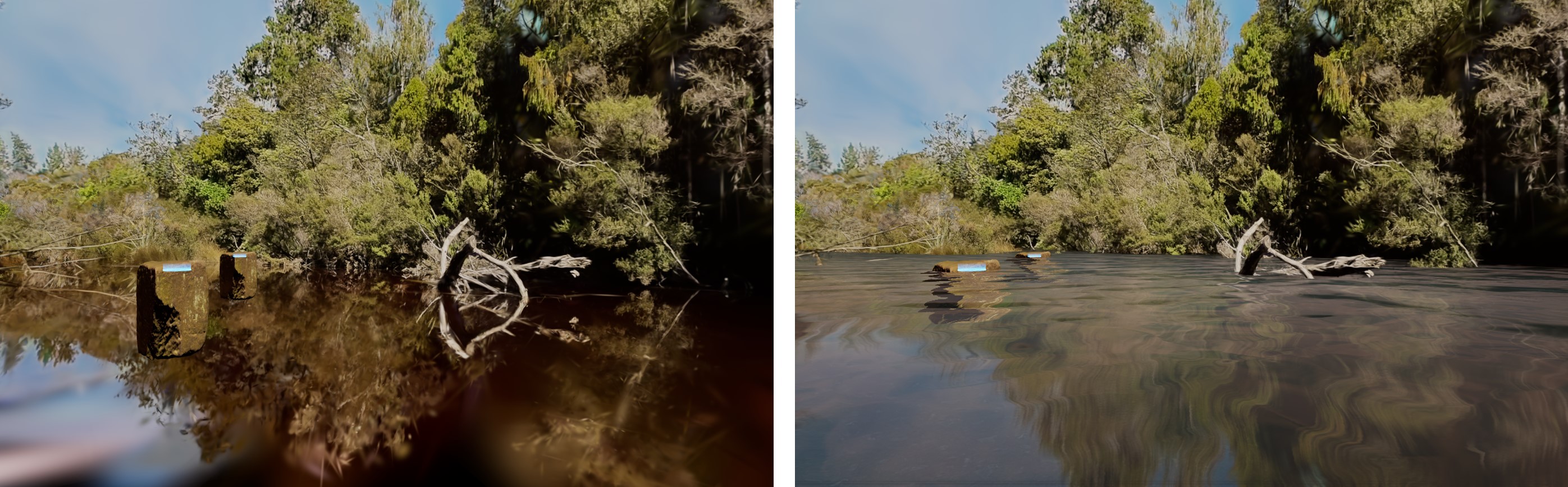}
    \caption{\textbf{Plain \ac{gs} (left) and our \approach (left).} For the main study we investigate the impact on adding 3D assets with dynamic characteristics (water, water current) and their impact on the user while moving outside the camera path.}
    \label{fig:conditions_main}
\end{figure*}

\subsection{Main Study}

Our main study investigates the effect of \approach in combination with 3D assets compared to plain \ac{gs}. In our preliminary study, we received feedback that self-directed movements would be appreciated ($N=10$). Thus, the user could now move freely in the virtual world by teleportation. We investigated whether \approach leads to a higher sense of presence when the user can move freely. We developed the following hypotheses based on previous indications and literature~\cite{slater1998influence,szita2018effects}: \\

~\textbf{HM1:} The addition of 3D assets into \ac{gs} using \approach will induce significantly higher spatial presence in users than plain \ac{gs}.
Given that the quality of \ac{gs} in terms of accurate reflections is decreasing with varying viewpoints, we assume that the 3D assets, i.e. water, can improve realism and sense of presence, as the reflections adapt to the viewpoint of the user.

~\textbf{HM2:}  We hypothesize that \approach is more graphically pleasing and visually coherent than plain \ac{gs}.
Considering, the relevance of the captured camera trajectory for \ac{gs} and \ac{nvs} in general, we expect a higher rate for visual coherence in \approach, as 3D assets have the potential to enrich the consistency of the overall 3D scene. 

\subsubsection{Study Setup}

\paragraph{Measures}

We measured presence with \ac{ipq}~\cite{schubert_sense_2003}. Participants again rated their favorite experience, commenting if wanted.

In comparison to the explorative study, we extended the personal preference questions by asking for graphical pleasing, visual coherence  and presence, as well as about the behavior of 3D environment, see~\autoref{tab:main_study_questions} for details. These were rated on a scale from 1 to 10, with 1 representing strong disagreement and 10 representing strong agreement. Inspired by Mal et al.~\cite{mal_virtual_2022}, we created these questions to analyze the perceived quality of the 3D environment.

\paragraph{Participants}
We recruited 30 participants (16 male, 14 female, 0 non-binary) who had no overlap with participants from the explorative study. The participants had an average age of 26.97$\pm$3.37).

\begin{table}[t!]
    \centering
    \caption{\textbf{Results of the \ac{ipq} from the main study.}}
    \resizebox{\columnwidth}{!}{
    \begin{tabular}{l|cccccc} \toprule
         \ac{ipq} &  $M_{GS}$ & $SD_{GS}$ & $M_{\approach}$ & $SD_{\approach}$  & $t(df)$ & $p$ \\ \midrule
         General  &  4.00 & 1.40 & 4.50 & 1.03  & -3.378(29) & \textbf{0.002} \\
         Spatial &  3.50 & 0.94 & 4.00 & 0.82 & -3.062(29) & \textbf{0.005} \\
         Involvement &  2.75 & 0.75 & 3.38 & 1.70 & -1.586(29) & 0.120   \\
         Realism &  1.63 & 0.90 & 2.75 & 0.84 & -6.755(29)  &$<$\textbf{0.001} \\ \midrule
         Overall &  2.64 & 0.90 & 3.47 & 0.86 & -4.015(29) & \textbf{0.007}\\ \bottomrule
    \end{tabular}}
    \label{tab:kayak_main_study}
\end{table}

\begin{table}[t!]
    \centering
    \caption{\textbf{Preference rating.} Averaged result of median $M$ and standard deviation $SD$ reported from the three locations in the virtual environment.}
    
    \resizebox{\columnwidth}{!}{
    \begin{tabular}{l|cccccc} \toprule
        & $M_{GS}$ & $SD_{GS}$ &  $M_{\approach}$ & $SD_{\approach}$ & $t(df) $& $p$  \\  \midrule 
        \textit{How present do you feel} \\
        ~~\textit{in the environment?} & 5.33 & 1.66 &6.58 & 1.42 & -4.016(29) &$<$\textbf{0.001}   \\
        \textit{How ... is this location} \\
        ~~\textit{... graphically pleasing ...} & 5.50 & 1.59 & {6.11} & {1.55} & -3.610(29) & \textbf{0.001} \\
        ~~\textit{... visually coherent ...} & 4.67 & 1.67 & {6.11} & {1.55} & -5.587(29) &$<$\textbf{0.001}\\
        \textit{The water was a } \\
        ~~\textit{ plausible part.} & 5.50 & 2.25 & {8.18} & {1.39} & -6.965 (29) & $<$\textbf{0.001} \\
        \textit{The reflection} \\
        ~~\textit{in the water matched.} & 5.50 & 2.40 & {7.98} & {1.40} & -6.781 (29) &  $<$\textbf{0.001}\\  \bottomrule
    \end{tabular}
    }
    \label{tab:main_study_questions}
\end{table}

\paragraph{Design}

We followed the same setup as in the explorative study, enriching the experience by allowing free choice of movement within a predefined space. For comparability between the participants, we selected three spots highlighted with rocks and teleportation indicators, see \autoref{fig:conditions_main}. As users could now move freely, we removed the $360^{\circ}$ video from the set of conditions and only compared \ac{gs} and \approach. We asked each participant to spend some time in \ac{vr}, exploring the surroundings before moving to each rocks with teleportation indication where they were asked to look around for 20 seconds each, before answering the questions. 

In contrast to the exploratory study, each user received a teleportation tutorial before the actual conditions began. We then followed the same procedure as in the exploratory study.

\begin{figure}[t!]
    \centering

    \includegraphics[width=0.9\columnwidth]{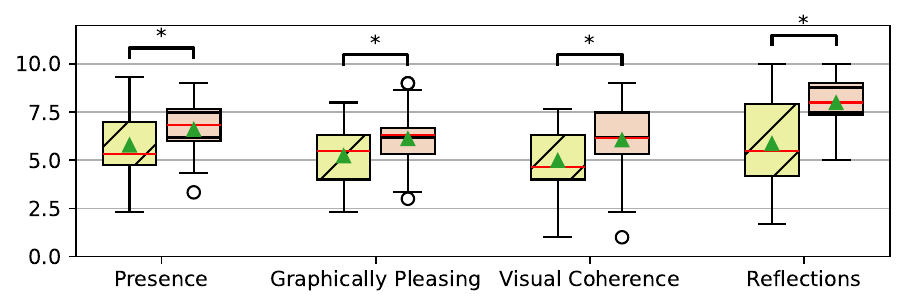}
    
    \caption{\textbf{Results of the individual preference questions from the selected spots.}}
    \label{fig:individual_questions}
\end{figure}

\begin{figure*}[t!]
    \centering
    \includegraphics[width=0.9\linewidth]{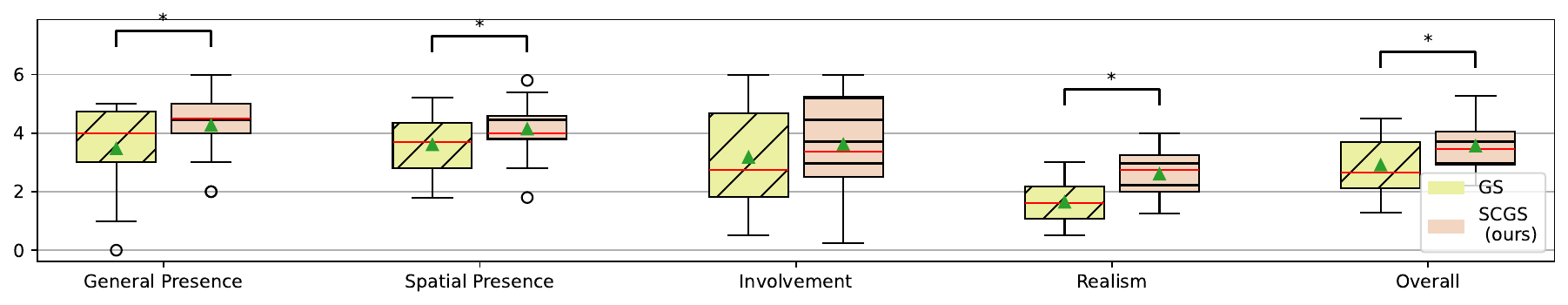}
    
    \caption{\textbf{\ac{ipq} results of the main study.} In the main study,  we measured a significant difference for general presence, spatial presence, realism, and overall presence.}
    \label{fig:ipq}
\end{figure*}

\subsubsection{Results and Discussion}

In our main study, participants experienced the same virtual environment but could now move freely. Given these new opportunities, we found a clear significant indication for General Presence, Spatial Presence, Realism and Overall Presence, clearly showing that participants felt more presence in the \approach generated scene, see \autoref{tab:kayak_main_study} and \autoref{fig:ipq}.

The overall ranking shows a clear preference for \approach, as 28 out of 30 participants preferred our approach. Looking at the individual questions for each condition's visual coherence, graphical pleasing or appearance of the reflections, we found that all aspects were ranked higher for \approach. Moreover, \approach significantly scored higher for presence, which is consistent with the \ac{ipq}. Furthermore, visual coherence and realism in terms of reflections also achieved a higher scoring when using \approach.

Several participants commented positively on \approach and the 3D asset (flowing water): \textit{``The movement of the water made the experience more realistic''} and \textit{``Water effects and spatial layout were very presence-provoking.''}. 

In terms of criticism, the water current and depth were mentioned by $N=4$:\textit{``I feel more realistic, but it would be better if the ground of the water gets deeper.''}. A comment possibly pointing to future work was: \textit{``It is an environment where sounds are expected, that felt like a reminder that it was not real.''}. This is consistent with feedback from the exploratory study. While the focus of our current work was on the visuals, spatial sound could be part of future work.

With regards to our two main hypotheses, a higher sense of presence is measured using \ac{ipq} when comparing plain \ac{gs} and \approach. We found significant differences in general presence, spatial presence, realism, and overall presence, confirming \textbf{HM1}, see \autoref{fig:ipq} and \autoref{tab:kayak_main_study}. At the individual spots participants were asked a general presence question related to the current location. There, we found no statistical significance, see \autoref{fig:individual_questions}. However, the mean and median show a higher indication as well as less standard deviation when using \approach. According to the evaluation on visual coherence and graphical pleasing, we can confirm \textbf{HM2}.

\section{General Discussion}

We propose \approach, a Semantics-Controlled Gaussian Splatting approach enabling 3D scene editing with large scale objects. Our approach is demonstrated on our proposed outdoor dataset and additional captures. Further evaluation for the segmentation quality on the 3D-OVS dataset shows that we are inline with the state-of-the-art. Additionally, we performed two evaluations that capture the experience of individual users.

\subsection{Technical Aspects} 

Our approach enables the segmentation and removal of large scene parts, outperforming the state-of-the-art in image quality and, as shown in \autoref{fig:removal}, in object removal.

Existing semantic 3D \ac{gs} approaches typically focus on scenes where a camera revolves around a single large object~\cite{knapitsch_tanks_2017} or multiple smaller objects~\cite{ye_gaussian_2023,liu_weakly_2023} but show limitations in their capability of removing large scene parts. As shown in \autoref{fig:removal} and \autoref{fig:usecasesexample}, our approach can not only handle smaller and larger objects, it is additionally capable of editing large scenes. This key element of our approach is enabled by directly assigning the class IDs to the Gaussians. With the adapted rasterization process, our approach can handle more diverse datasets. 

To validate our approach, we propose a rather complex dataset, capturing large outdoor scenes with a path-following setup. The dataset is captured in a different setup compared to existing work~\cite{ye_gaussian_2023} posing new challenges to separable \ac{gs}. The forward-motion of the camera in out dataset results in a few frames per spot, challenging both \ac{gs} approaches as well as preprocessing. As depicted in \autoref{fig:removal}, \approach can better handle this new dataset and is able to remove parts of the scene without affecting remaining parts. Moreover, as shown in \autoref{tab:rendering_result_our_data}, our approach leads to improvements in \ac{nvs} quality on this dataset.

\subsection{User Evaluation}

Participants generally responded positively to \approach, particularly when they were allowed to move freely in the outdoor surroundings. When tied to the camera path, users preferred the original video which is conclusive with previous research~\cite{teo_mixed_2019} on other 3D reconstruction approaches. In our main study, we found significant differences for enhanced realism and presence in the scene generated with \approach compared to plain \ac{gs}. Criticism of the \approach-generated scene focused on the water's depth and current, with $N=4$ participants suggesting improvements.
Our findings support our hypotheses:

\emph{\textbf{HM1:} The addition of 3D assets into \ac{gs} using \approach will induce significantly higher spatial presence in users than \ac{gs} alone.}.

\emph{\textbf{HM2:}  We hypothesize that \approach is more graphically pleasing and visually coherent than plain \ac{gs}}.

The questions asked at the individual locations within the scenery and the \ac{ipq} confirmed that with free user interaction, \approach shows advantages in realism, visual quality, visual coherence and presence when directly compared to plain \ac{gs}. The positive perception of \approach was further supported by the preference ratings. 28 out of 30 participants preferred \approach over plain \ac{gs}. Those who preferred plain \ac{gs} mentioned that they appreciated the stillness of the scene.

\subsection{Limitations}

From a technical perspective, our approach is strongly dependent on predefined labels. Therefore, new scenes with inconsistent labels are challenging for our approach as we directly assign the labels.

Our study investigates the advantage of using our large-scale scene parts together with 3D assets from a Game Engine. A large, regularly dynamic part is replaced by a 3D asset. We assume that when parts of a 3D scene that are less influenced by the environment are replaced, e.g., a car or concrete of the street, the effects in presence or preference could be lower.

\subsection{Future Work}

From a technical and user perspective, future work could look at floating splats far outside the camera path where the 3D position is not accurately learned. Removing these could be beneficial for users who move freely. Moreover, our dataset offers potential for further improvements, for example, object removal and \ac{nvs} quality for large outdoor scenes.

In addition, future work for \ac{vr} could integrate more targeted user interactions, such as rowing~\cite{hedlund_rowing_2024,shoib_rowing_2020,keller_augmenting_2021}, if the environment contains water, or walk-in-place, if the environment includes hiking areas~\cite{haliburton_vr-hiking_2023,alvarado_trail_2024}. 
As mentioned by the participants, sound would be beneficial for a more realistic experience. For comparability reasons, we have deliberately limited our study to the visual representation and have intentionally omitted the sound, as sound can have an effect on presence~\cite{9098089,10322289}.

\section{Conclusion}
Overall, we present a novel approach for 3D asset generation based on Semantics-Controlled \ac{gs}, alongside a new dataset featuring challenging outdoor scenes that pose various difficulties for \ac{nvs}. 

In summary, \approach introduces an enhanced \ac{gs} approach for generating large-scale 3D assets in \ac{vr}. We evaluated our method from both a technical and user perspective. In the user study, we set a baseline for presence on our scenes. Therein, \approach was compared to plain \ac{gs}, with results demonstrating that \approach significantly outperforms plain \ac{gs} in terms of presence and perceived quality when users move freely within the environment. From a technical perspective, we outperform the state-of-the-art in object removal and scene editing on our new dataset. For segmentation quality we provide state-of-the-art results demonstrating that \approach can handle a variety of different scenes. Additionally, we showcased our approach for other use cases outside of its purposed dataset, showing promising results fostering \ac{vr} research.

\section*{Acknowledgments}
The authors acknowledge funding by DAAD for research stays of computer scientists across the world.

The authors gratefully acknowledge the scientific support and HPC resources provided by the Erlangen National High Performance Computing Center (NHR@FAU) of the Friedrich-Alexander-Universität Erlangen-Nürnberg (FAU).

\bibliographystyle{abbrv}
\bibliography{references}

\end{document}